\documentclass{iopjournal}

\usepackage{amsmath,amssymb}
\usepackage{array,tabularx}
\usepackage{enumitem}
\usepackage{float}
\usepackage{algorithm}
\usepackage{algpseudocode}
\usepackage{microtype}
\usepackage{url}
\usepackage{indentfirst}

\graphicspath{{figures/}}

\newcolumntype{L}[1]{>{\raggedright\arraybackslash}p{#1}}
\newcolumntype{C}[1]{>{\centering\arraybackslash}p{#1}}

\makeatletter
\renewcommand\subsubsection{\@startsection{subsubsection}{3}{0pt}%
  {-2.0ex plus -0.6ex minus -0.2ex}%
  {0.65ex plus 0.2ex}%
  {\normalfont\normalsize\itshape\raggedright}}
\makeatother

\begin{document}

\articletype{Paper}

\title{MIRA-PRM: Mission-Informed Reusable Roadmap Planning for Mobile Gas-Sensing Inspection}

\author{Gongsen Wang$^{1,2}$, Siyuan Wang$^{3,*}$, Xinyuan Wang$^4$, Yuhan Wen$^5$, Feng Yang$^4$, Zhen Tian$^6$ and Hoi Leong Lee$^{7,*}$ }

\affil{$^1$State Key Laboratory of Transducer Technology, Aerospace Information Research Institute, Chinese Academy of Sciences, Beijing 100190, China}
\affil{$^2$School of Electronic, Electrical and Communication Engineering, University of Chinese Academy of Sciences, Beijing 100049, China}
\affil{$^3$School of Electronic and Optical Engineering,Zijin College,Nanjing University of Science and Technology,Nanjing 210023, China}
\affil{$^4$School of Electrical and Automation Engineering, Nanjing Normal University, Nanjing 210023, China}
\affil{$^5$School of Information and Communication Engineering,North University of China, Taiyuan 030051, Shanxi, China}
\affil{$^6$School of Computing Science, University of Glasgow, Glasgow G12 8QQ, UK}
\affil{$^7$Faculty of Electronic Engineering and Technology, Universiti Malaysia Perlis, Arau 02600, Malaysia}
\affil{$^*$Corresponding authors: Siyuan Wang and Hoi Leong Lee.}

\email{Siyuan Wang: 19852225218@163.com; Hoi Leong Lee: hoileong@unimap.edu.my}


\begin{abstract}
Mobile gas-sensing inspection depends not only on sensor performance but also on whether the mobile platform can reliably reach scheduled sampling poses. This work addresses that route-availability requirement through MIRA-PRM, a mission-informed probabilistic roadmap that combines geometry-, mission-flow-, and inspection-conditioned sampling with locally adaptive connectivity and evidence-gated refinement. Eight development campaigns comprising 142,448 planner runs characterized mission scaling, repeated-query adaptation, finite resident-node budgets, target distributions, local map repair, mechanism ablation, cross-family comparison, and parameter sensitivity. MIRA-PRM maintained 100\% seed-level success as the number of ordered targets increased from two to eight, while mean planning time increased from 0.0570 to 0.0694 s. Under a 40-node ceiling in the bottleneck campaign, seed-level success was 97.67\%, compared with 67.17\% for PRM and 86.50\% for PRM*. Local repair reduced event time by 79.9--81.9\% and node additions by 90.8--94.3\% relative to full rebuilding, with a small path-quality penalty. A separate frozen holdout contained 12 procedurally generated maps, 48 ordered missions, and three planner seeds per map--mission unit. MIRA-PRM produced 96/144 validated successes, compared with 82/144 for PRM, 89/144 for PRM*, and 70/144 for LE-HG-PRM. Among paired runs in which both planners succeeded, MIRA-PRM reduced wall time by 16.89\% relative to PRM* and by 81.01\% relative to the contextual per-query LE-HG-PRM implementation, but remained 52.23\% slower than PRM; its paths were 0.54--1.86\% longer. Every validated path passed independent full-segment checks and satisfied the clearance non-inferiority criterion. Because the frozen manifest contained three rather than the protocol-specified five seeds, the prespecified 5/5 robust-success endpoint could not be evaluated. The results support a reliability-oriented planning trade-off in structured internal simulations and motivate external-map and physical mobile-sensing validation.
\end{abstract}

\setlength{\parindent}{2em}

\section{Introduction}

Mobile robots extend gas sensing beyond fixed monitoring stations by carrying a common sensing payload through spatially distributed sampling locations, thereby increasing spatial coverage and reducing human exposure in hazardous or inaccessible environments \cite{ref1,ref2}. Portable analytical techniques such as ion mobility spectrometry (IMS) further support trace-level volatile-organic-compound measurements outside conventional laboratory workflows \cite{ref3}. In a mobile measurement system, however, analytical performance is only one part of data availability: if the platform cannot reach a prescribed sampling pose, the corresponding observation is absent before the sensing chain can operate. Route completion is therefore an upstream availability requirement for repeated mobile gas measurement.

Repeated inspection differs from a single start--goal planning query because the platform must traverse an ordered sequence of sampling poses and can reuse the same doors, aisles, and constrained passages over multiple mission legs. This repeated structure makes the allocation of a finite roadmap particularly important. Classical probabilistic roadmaps (PRMs) construct reusable graphs from free-space samples \cite{ref4}, but finite uniform sampling can underrepresent gates and narrow passages. PRM* changes connectivity as the roadmap grows to obtain asymptotic optimality properties \cite{ref5}, while Lazy PRM postpones collision checking to reduce unnecessary edge evaluation \cite{ref6}. These methods do not, by themselves, allocate limited roadmap capacity according to the ordered measurement mission.

A number of recent planners use environmental structure or search-state information to improve sampling efficiency. HG-PRM introduced heuristic-guided sampling and dynamic weighting, and LE-HG-PRM subsequently incorporated explicit environmental structure into roadmap construction \cite{ref7,ref8}. MMD-RRT adapts tree sampling and expansion to local obstacle density \cite{ref9}; hybrid ACO--RRT methods use an initial tree solution to guide subsequent optimization \cite{ref10}; and global--local or artificial-potential-field strategies provide complementary mechanisms for dynamic conflict resolution and local-minimum avoidance \cite{ref11,ref12}. These approaches demonstrate the value of incorporating environmental and search-state information, but they do not address how a finite persistent roadmap should be allocated and subsequently refined according to an ordered sensing mission. This is the specific gap considered here.

We therefore develop the Mission-Informed Reusable Adaptive Probabilistic Roadmap (MIRA-PRM). MIRA-PRM combines probability-mass priors derived from constrained geometry, ordered mission flow, and inspection-pose support with locally adaptive connectivity and evidence-gated graph growth. The method does not treat roadmap reuse as a new PRM concept; instead, it uses persistence as the substrate on which mission-specific sampling and bounded, evidence-aligned refinement operate.

The contributions are fourfold.

\begin{enumerate}[label=(\arabic*),leftmargin=2.3em]
\item A mission-conditioned probability model combines constrained geometry, ordered route demand, and inspection-pose support after independent normalization of the active prior fields. This gives the sampling weights a consistent probability-mass interpretation and concentrates finite roadmap capacity in mission-relevant free space.

\item A persistent mission-level roadmap couples locally adaptive connectivity with evidence-gated refinement. Disconnection, insufficient guide support, detour, and path--guide deviation can trigger spatially targeted growth, while marginal-improvement stopping, cooldown, and global/segment budgets prevent unconditional densification.

\item Eight full-scale development campaigns characterize scaling, repeated-query behaviour, finite-budget performance, target-distribution dependence, dynamic repair, mechanism ablation, cross-family behaviour under configured settings, and one-factor parameter sensitivity. These campaigns are treated as mechanism and development evidence rather than as independent population-level validation.

\item A separate frozen paired holdout evaluates validated route completion, end-to-end planning time, failure-aware PAR-2, path length, full-segment clearance, and resident-node compliance. The holdout is analysed separately from development data so that tuning evidence and frozen evaluation are not conflated.
\end{enumerate}

\section{Problem formulation and evaluation}

\subsection{Ordered gas-sensing inspection route}

Let \(G\) be an \(H\times W\) binary occupancy grid, with \(G(x)=1\) for free space and \(G(x)=0\) for obstacles, and let \(F=\{x:G(x)=1\}\). An inspection mission is an ordered sequence \(Q=(q_0,q_1,\ldots,q_m)\), where \(q_0\) is the deployment pose and \(q_1,\ldots,q_m\) are prescribed gas-sampling poses. For every consecutive pair \((q_{j-1},q_j)\), the planner must provide a collision-free path \(\pi_j\). The visit order is determined upstream by the measurement schedule; MIRA-PRM solves the geometric routing layer and does not optimize or reorder the sampling sequence.

\begin{equation}
\begin{aligned}
\Pi&=\{\pi_1,\ldots,\pi_m\},\qquad
\pi_j:q_{j-1}\rightarrow q_j,\\
&\hspace{5.3em}\pi_j\subset F.
\end{aligned}
\label{eq:mission-paths}
\end{equation}

A mission is regarded as validated only if every required leg is completed before the common deadline, the prescribed endpoints are reached in order, the applicable resident-node limit is respected, and the delivered geometry passes an independent full-segment collision check. This definition is intentionally mission-level: failure of one leg removes at least one scheduled measurement opportunity.

\begin{figure}[t]
\centering
\includegraphics[width=0.95\linewidth]{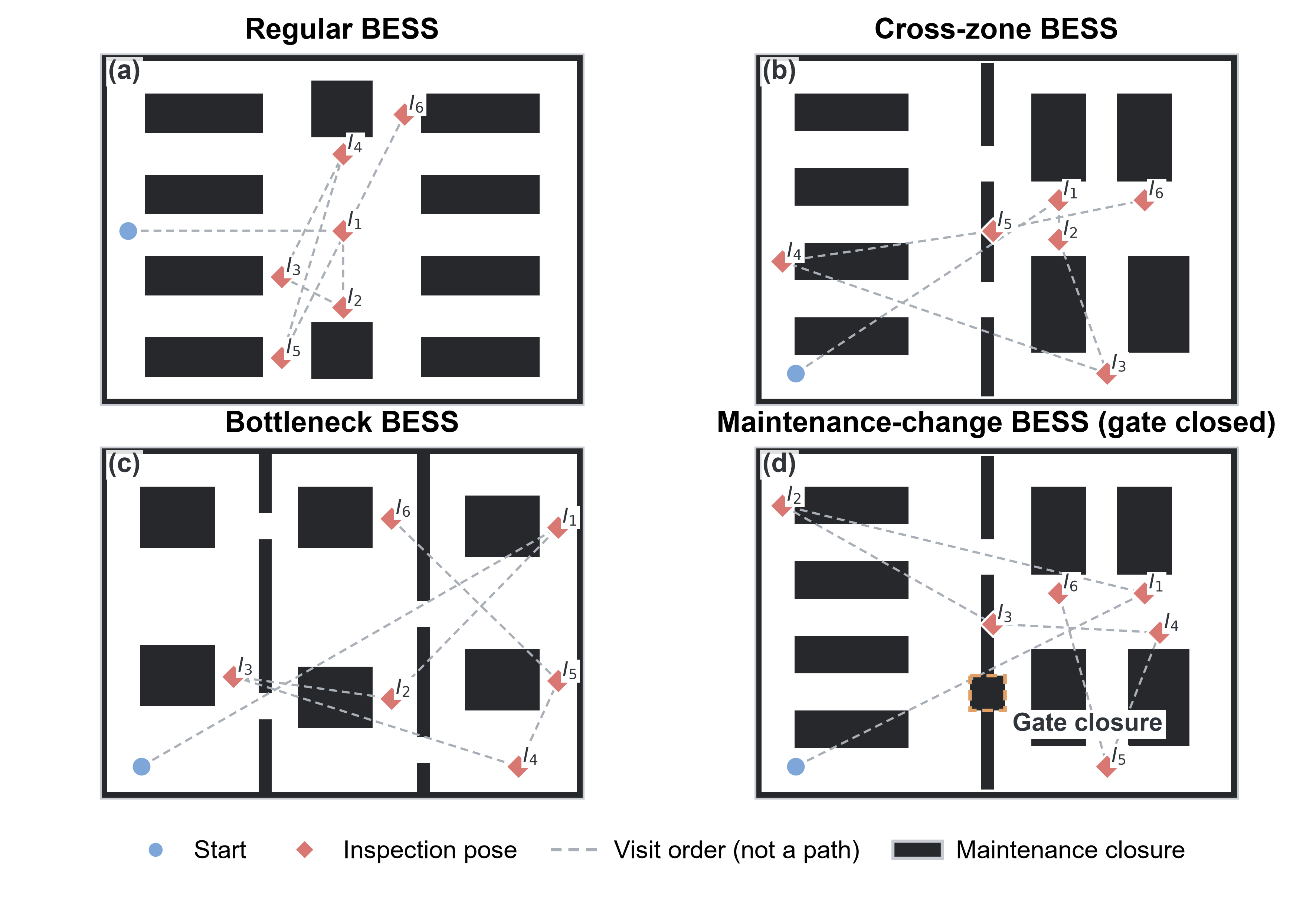}
\caption{BESS-inspired structured industrial layouts used in the development experiments. The geometries represent regular rows, cross-zone traversal, a bottleneck, and temporary gate closures. Blue circles denote the deployment pose, coral diamonds denote ordered inspection poses, and dashed segments indicate the prescribed visit order rather than executable paths. These layouts support the Exp.1--8 development campaigns; the frozen holdout uses a separate set of procedurally generated maps.}
\label{fig:environments}
\end{figure}

\subsection{Routing performance quantities}

Four outcome families are reported. Validated mission success is the primary route-availability quantity. End-to-end wall time includes planner initialization, prior construction, roadmap construction, all mission queries, and the common path-transforming post-processing applied inside the planning workflow. Path length is compared on pairwise common-success cases so that geometric effects are not confounded by missing paths. Geometric safety is checked on the delivered path using a supercover traversal of every segment together with a Euclidean distance transform \cite{ref13}.

Failures are retained in the efficiency analysis through PAR-2 with a deadline \(T=60\) s:
\begin{equation}
t_i^{\mathrm{PAR\text{-}2}}=
\begin{cases}
t_i, & \text{if run }i\text{ is validated},\\
2T, & \text{otherwise}.
\end{cases}
\label{eq:par2}
\end{equation}

For pairwise comparison with baseline \(b\), positive relative effects denote lower MIRA-PRM time or path length:
\begin{equation}
E_{\mathrm{time}}=100\frac{t_b-t_{\mathrm{MIRA}}}{t_b},\qquad
E_{\mathrm{length}}=100\frac{L_b-L_{\mathrm{MIRA}}}{L_b}.
\label{eq:relative-effects}
\end{equation}

\subsection{Measurement interpretation and uncertainty boundary}

The quantities above characterize the routing layer of a mobile measurement system; they are not estimates of gas-concentration uncertainty, detector accuracy, or analytical sensitivity. Statistical intervals in this study describe variation associated with the finite map, mission, and planner-seed samples. In a physical gas-sensing platform, additional uncertainty contributions would arise from localization, robot footprint and tracking error, sampling-pose tolerance, detector response time, calibration, and environmental transport. Those terms are outside the present simulation study and are therefore not combined with the routing statistics reported here.

\section{MIRA-PRM}

\subsection{Mission-conditioned probability-mass priors}

MIRA-PRM constructs three non-negative score fields over free space. The geometry field \(g\) emphasizes traversable bands near structured obstacles, where finite uniform sampling is most likely to miss topologically important passages. The mission-flow field \(f_m\) is accumulated along conservative coarse-grid guides connecting consecutive inspection poses, so corridors demanded by multiple mission legs receive additional support. The inspection field \(f_i\) places clearance-adaptive support around required sampling poses and is softly suppressed where mission-flow support is already high, reducing redundant concentration around open targets.

Each active field is converted independently to a probability mass function over \(F\). Inactive or zero-mass priors are omitted and the remaining prior weights are renormalized. With a uniform exploration fraction \(\epsilon\) and active weights \(w_g,w_m,w_i\), the sampling distribution is
\begin{equation}
p(x)=\frac{\epsilon}{|F|}+(1-\epsilon)
\left[w_g p_g(x)+w_m p_m(x)+w_i p_i(x)\right],\qquad x\in F,
\label{eq:sampling-density}
\end{equation}
where the active non-uniform weights sum to one. The principal evaluation profile uses \(\epsilon=0.15\) and \((w_g,w_m,w_i)=(0.24,0.54,0.22)\). Independent normalization is important because these values then represent allocated probability mass rather than arbitrary peak amplitudes.

\begin{figure}[t]
\centering
\includegraphics[width=0.95\linewidth]{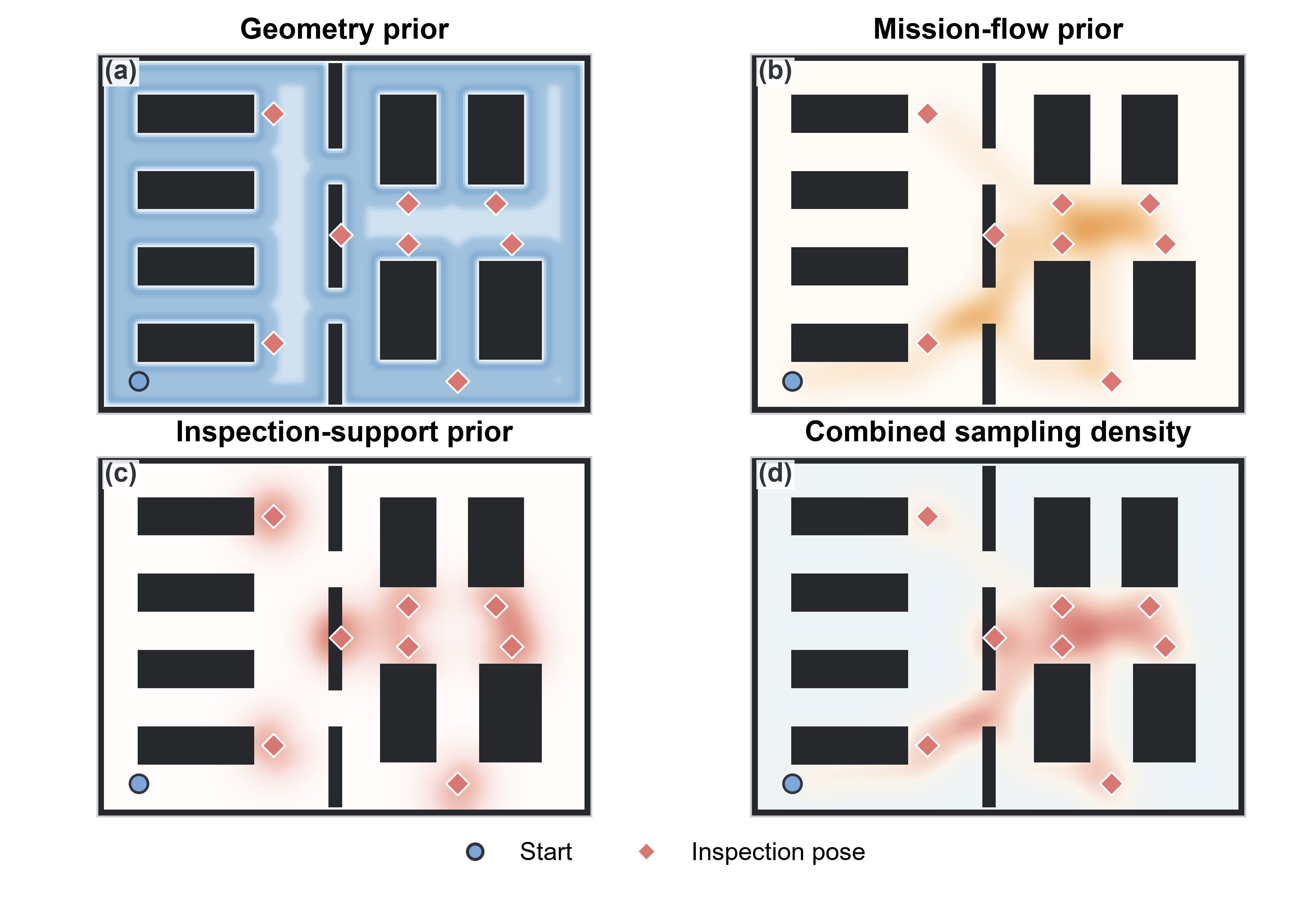}
\caption{Construction of the MIRA-PRM sampling distribution. Geometry support emphasizes constrained free-space bands; mission-flow support follows conservative guides between ordered sampling poses; inspection support concentrates around required poses; and the final density combines the active normalized fields with a 15\% uniform exploration component. Intensity indicates relative sampling density, not the probability that a path follows a particular corridor.}
\label{fig:sampling-priors}
\end{figure}

\subsection{Persistent roadmap and adaptive connectivity}

All mission endpoints are reserved during sampling, and a single roadmap is maintained across the ordered mission. The initial roadmap population \(N_0\) is a configurable algorithm parameter rather than a universal fixed fraction of a node ceiling. In the matched-resident frozen experiments only, the resource protocol sets
\[
N_0=\max\left\{|\operatorname{Unique}(Q)|,\left\lceil0.40N_{\max}\right\rceil\right\},
\]
leaving the remainder of \(N_{\max}\) available for refinement. Other experiments use their stated configured initial population and growth limits.

Candidate edges are retained only if their corresponding grid segments are collision-free. Connectivity is increased locally in constrained regions according to the normalized geometry score:
\begin{equation}
k(x)=
\begin{cases}
\lfloor1.60k_0\rfloor, & g(x)>0.80,\\
\lfloor1.30k_0\rfloor, & 0.55<g(x)\leq0.80,\\
k_0, & \text{otherwise}.
\end{cases}
\label{eq:adaptive-connectivity}
\end{equation}
For the frozen holdout profile, \(k_0=5\), yielding 8 and 6 neighbours in the high- and intermediate-geometry bands, respectively. This local increase compensates for the higher probability of collision-rejected neighbours near structured obstacles without globally densifying the roadmap. Once endpoints are present, graph A* is used for the route query \cite{ref14}.

\subsection{Evidence-gated refinement}

After each roadmap query, MIRA-PRM evaluates whether the current graph provides adequate support for the requested transition. The diagnostic is scale-normalized using the characteristic initial spacing
\begin{equation}
h=\sqrt{\frac{|F|}{N_0}},
\label{eq:support-scale}
\end{equation}
and combines guide-to-roadmap support, the fraction of unsupported guide samples, path--guide deviation, detour, and search load. The initial population is used in \(h\) so that the trigger does not become progressively stricter merely because refinement has already added nodes.

A disconnected query or severe support deficit is immediately eligible for refinement. A successful first visit uses a higher trigger threshold unless support is severe, limiting the tendency to spend persistent capacity on one-off improvements. If refinement is enabled, spatial evidence determines where new support is placed: disconnection maps to deterministic bridge anchors, support gaps to sparse guide regions, path--guide deviation to separated guide regions, and detour evidence to distributed guide support. Search load contributes to the trigger score but is not used as an independent spatial target. Growth is bounded by a global feedback cap, a per-query round limit, a per-segment batch limit, and a cooldown activated when successive refinements provide insufficient marginal improvement.

\begin{table}[t]
\caption{MIRA-PRM mechanisms and their functional roles.}
\label{tab:mechanisms}
\centering
\small
\begingroup
\renewcommand{\arraystretch}{1.12}
\begin{tabularx}{\linewidth}{@{}L{0.20\linewidth} L{0.31\linewidth} X@{}}
\hline
\textbf{Component} & \textbf{Information used} & \textbf{Functional role} \\
\hline
Geometry prior & Constrained free-space bands & Improves representation of gates and narrow passages. \\
Mission-flow prior & Ordered guide paths across all mission legs & Allocates persistent capacity to repeatedly demanded corridors. \\
Inspection support & Clearance-adaptive support around required poses & Supports endpoint connectivity without duplicating strong mission-flow mass. \\
Adaptive connectivity & Larger local \(k\) in constrained regions & Compensates locally for collision-rejected neighbours. \\
Evidence-gated refinement & Disconnection, support, detour and deviation diagnostics & Adds nodes only where route-level evidence indicates insufficient graph support. \\
\hline
\end{tabularx}
\endgroup
\end{table}

\subsection{Algorithm}

Algorithm~\ref{alg:mira} summarizes the nominal construction and ordered-query loop. The prior normalization, refinement gate, cooldown logic, bounded growth, and evidence-aligned insertion steps are shown explicitly, while low-level diagnostic arithmetic is omitted for clarity. The map-update local-repair routine evaluated in Exp.5 is a separate operation. Full-resolution reference A*, independent full-segment collision/clearance validation, metric evaluation, and statistical analysis are performed outside the timed planner.

\begin{algorithm}[H]
\caption{MIRA-PRM for ordered mobile gas-sensing inspection}
\label{alg:mira}
\footnotesize
\begin{algorithmic}[1]
\Statex \textbf{Input:} occupancy grid \(G\); ordered poses \(Q=[q_0,\ldots,q_m]\); parameters \(\theta=(N_0,k_0,R_{\max},B_{\mathrm{seg}},B_{\max})\)
\Statex \textbf{Output:} collision-free segment paths \(\Pi\); persistent roadmap \(RM\); segment diagnostics \(D\)
\State \(F\gets\{x:G(x)=1\}\); \(C\gets\Call{EuclideanDistanceTransform}{F}\)
\State \((g,f_m,f_i,\mathrm{Guides})\gets\Call{BuildPriorsAndGuides}{G,Q,C,\theta}\)
\State \(p_g,p_m,p_i\gets\operatorname{PMF}_F(g),\operatorname{PMF}_F(f_m),\operatorname{PMF}_F(f_i)\)
\State \(p\gets\varepsilon U_F+(1-\varepsilon)(w_gp_g+w_mp_m+w_ip_i)\), \(w_g+w_m+w_i=1\)
\State \(U\gets\operatorname{Unique}(Q)\); \(V\gets U\cup\Call{WeightedSampleWOR}{p,N_0-|U|;F\setminus U}\)
\State \(RM\gets\Call{BuildCollisionFreeAdaptiveRoadmap}{G,V,g,k_0}\); \(h\gets\sqrt{|F|/|V|}\)
\State \(\Pi,D\gets\langle\rangle,\langle\rangle\); \Call{Initialize}{budgets,state}
\For{each consecutive mission segment \((s,t)\) in \(Q\)}
    \State \Call{EnsureEndpoints}{$RM,s,t$}; \(st\gets\mathrm{State}[\operatorname{UnorderedPair}(s,t)]\)
    \State \(cool\gets\Call{BeginSegmentVisit}{st}\); \(prev\gets\varnothing\)
    \For{\(r\gets0,\ldots,R_{\max}\)}
        \State \(res\gets\Call{RoadmapAStar}{RM,s,t}\); \(d\gets\Call{Diagnose}{res,\mathrm{Guides}(s,t),RM,h,\theta}\)
        \If{\(prev\neq\varnothing\land\Call{BothMarginalGainsBelow}{prev,(res,d);\eta_L,\eta_S}\)}
            \State \Call{SetCooldown}{st}; \textbf{break}
        \EndIf
        \State \(first\gets(st.\mathrm{visits}=1\land res.\mathrm{success})\); \(severe\gets(d.\mathrm{support\_excess}\ge\zeta)\)
        \State \(\tau_{\mathrm{eff}}\gets\tau_0+\mathbf{1}\{first\land\neg severe\}\Delta\tau\)
        \State \(gate\gets(\neg res.\mathrm{success})\lor severe\lor(d.\mathrm{score}\ge\tau_{\mathrm{eff}})\)
        \State \(B_{\mathrm{rem}}\gets\Call{RemainingFeedbackBudget}{budgets,B_{\max},|RM|}\)
        \State \(eligible\gets gate\land(r<R_{\max})\land(B_{\mathrm{rem}}>0)\)
        \Statex \hspace{\algorithmicindent}\(\phantom{eligible\gets{}}\land(st.\mathrm{batches}<B_{\mathrm{seg}})\land(\neg cool\lor\neg res.\mathrm{success})\)
        \State \textbf{if} \(\neg eligible\) \textbf{then break}
        \State \(prev\gets(res,d)\); \(X\gets\Call{EvidenceAlignedRefine}{d,B_{\mathrm{rem}},\theta}\)
        \State \textbf{if} \(X=\varnothing\) \textbf{then break}
        \State \Call{InsertAndAdaptiveConnect}{$RM,X,g,k_0$}; \(st.\mathrm{batches}\gets st.\mathrm{batches}+1\)
    \EndFor
    \If{\(res.\mathrm{success}\)}
        \State \(\pi\gets\Call{CollisionCheckedGreedyShortcut}{res.\mathrm{path},G}\); \Call{Append}{$\Pi,\pi$}
    \EndIf
    \State \Call{Append}{$D,\operatorname{SegmentDiagnostics}(res,d,st,res.\mathrm{success})$}
\EndFor
\State \Return \(\Pi,RM,D\)
\end{algorithmic}

\vspace{0.2em}
\scriptsize\textit{Note.}
\(U_F(x)=1/|F|\) on \(F\); each active prior is normalized as \(\operatorname{PMF}_F(a)(x)=a(x)/\sum_{y\in F}a(y)\), with inactive priors omitted and active weights renormalized. Coarse-grid guides support prior construction, diagnosis, and refinement targeting only. In the matched-resident frozen protocol, \(N_0=\max\{|\operatorname{Unique}(Q)|,\lceil0.40N_{\max}\rceil\}\), and feedback is additionally bounded by \(N_{\max}\). Reference A*, independent validation, metric evaluation, and statistical analysis are performed outside the timed planner.
\end{algorithm}

\section{Experimental design}

\subsection{Development and mechanism campaign}

Eight development campaigns were run using the full-scale experimental configuration to characterize scaling, mechanisms, sensitivity, and cross-family behaviour before the frozen holdout was examined. Together they contain 142,448 planner runs. These records are analysed only within their specified development maps and configurations and are not pooled with the frozen dataset.

Exp.1 tested 2, 4, 6, and 8 ordered targets on regular, cross-zone, and bottleneck maps, with 50 missions, 12 planner seeds, and seven algorithms in each map--target-count condition (50,400 records). Exp.2 used 48 repeated queries, arranged as six epochs of eight queries, for 12 missions, 12 seeds, and four planner variants (27,648 records). Exp.3 varied the resident-node ceiling over 40, 70, 100, 150, 250, and 400 nodes for 50 eight-target bottleneck missions, 12 seeds, and seven algorithms (25,200 records). Exp.4 tested clustered, distributed, cross-zone, and alternating target distributions with 50 missions per pattern, 12 seeds, and six algorithms (14,400 records).

Exp.5 compared local repair with complete roadmap rebuilding under gate-lower, gate-upper, and PCS-aisle closures, using 20 missions, eight seeds, and eight query stages per closure (7,680 records). Exp.6 combined a cumulative mechanism sequence with a leave-one-out ablation on 50 nine-target bottleneck missions and 12 seeds (6,000 records). Exp.7 compared eleven configured planner families on regular, cross-zone, and bottleneck maps with 30 eight-target missions and eight seeds per map (7,920 records). Exp.8 varied the uniform exploration fraction, feedback threshold, mission-flow probability mass, and refinement budget one factor at a time, using five levels per factor, 20 missions, and eight seeds (3,200 records).

\subsection{Frozen holdout evaluation (Exp.9)}

Exp.9 uses a frozen internal holdout generated independently of the Exp.1--8 development figures and experiments. Three topology families---maze, rooms, and random obstacles---each contribute four \(96\times128\) maps. Every map contains four fixed mission patterns (cross-zone, distributed, clustered, and random), six ordered inspection targets, and three planner seeds. The core campaign therefore comprises 12 maps, 48 map--mission units, and 576 planner runs. A complementary resource subset evaluates resident ceilings of 72, 120, and 168 nodes. The compared algorithms share the occupancy grids, mission order, collision model, and greedy path post-processor. For the resident-matched MIRA-PRM runs, the initial roadmap is set to 40\% of the resident ceiling (subject to endpoint reservation), with the remaining capacity available to evidence-gated refinement.

\begin{table}[t]
\caption{Frozen evaluation design.}
\label{tab:frozen-design}
\centering
\small
\begingroup
\renewcommand{\arraystretch}{1.10}
\begin{tabularx}{\linewidth}{@{}L{0.31\linewidth} X@{}}
\hline
\textbf{Design item} & \textbf{Value} \\
\hline
Topology families & Maze; rooms; random obstacles \\
Core maps / missions & 12 maps / 48 ordered missions \\
Inspection targets & 6 per mission \\
Planner seeds & 3 per map--mission unit \\
Core algorithms & MIRA-PRM; PRM; PRM*; LE-HG-PRM \\
Core resident ceiling & 120 nodes \\
Resource ceilings & 72; 120; 168 nodes \\
Timeout / PAR-2 & 60 s / 120 s \\
Uncertainty & 20,000 map-cluster bootstrap resamples; 95\% CI \\
Multiplicity & Holm adjustment within endpoint family \\
\hline
\end{tabularx}
\endgroup
\end{table}

\subsection{Comparison algorithms and evaluation protocol}

A* \cite{ref14} provides a deterministic grid-search reference under the common collision model and post-processing pipeline, allowing the sampling-based planners to be compared with a direct search baseline.

PRM \cite{ref4} is the classical reusable-roadmap comparator, selected to test whether mission-informed sampling improves on uniform sampling with a fixed neighbour count.

PRM* increases the neighbour count with roadmap size according to the k-nearest asymptotic-optimality principle \cite{ref5}, providing a comparator for the effect of a more strongly connected uniform roadmap.

PRM-Rebuild constructs a new PRM for each mission segment and serves as an execution-mode control for quantifying the benefit of roadmap reuse.

HG-PRM \cite{ref7} represents heuristic-guided roadmap construction and is included as the predecessor to the structure-aware LE-HG-PRM formulation.

LE-HG-PRM \cite{ref8} is implemented as a two-dimensional adaptation of the published structure-aware method and is selected as a contextual comparator for planning in structured environments; its graph is rebuilt for each query.

RRT represents single-query sampling-tree planning and is included to compare reusable roadmaps with incremental tree exploration.

RRT* \cite{ref5} adds neighbourhood-based parent selection and rewiring to tree exploration, providing a path-quality-oriented counterpart to RRT.

IAPF \cite{ref12} represents potential-field planning with local-minimum escape and adaptive steps in the implemented two-dimensional setting, broadening the comparison beyond roadmap and tree methods.

MMD-RRT \cite{ref9} combines multi-mode sampling with obstacle-density-adaptive steps in the implemented two-dimensional setting and is included as an adaptive tree-based comparator.

All methods use the same collision model and common path post-processing. Exp.7 compares the configured profiles of the ten comparators above with MIRA-PRM. Because graph budgets, iteration limits, collision-check semantics, and termination rules differ across planner families, Exp.7 is a profile comparison rather than a matched-computation ranking. The frozen holdout compares MIRA-PRM with PRM, PRM*, and LE-HG-PRM. PRM and PRM* share the resident-node ceiling with MIRA-PRM; LE-HG-PRM remains a contextual per-query comparator.

\subsection{Statistical analysis}

All recorded runs, including failures, were retained. For Exp.1--8, the analysis unit was a mission within its exact experimental condition. Planner seeds were first aggregated within mission so that repeated stochastic seeds were not treated as independent mission replicates. Robust success required every configured seed for a mission to succeed. Continuous endpoints used only mission pairs with finite values for both methods; path-quality comparisons additionally required finite path measurements under all configured seeds. Paired mission-level differences were evaluated with two-sided Wilcoxon signed-rank tests, with paired \(t\)-tests reported as sensitivity analyses. Mean-difference confidence intervals used 10,000 paired mission bootstrap resamples, and Holm adjustment was applied within each experiment and endpoint family \cite{ref19,ref20}. Exp.2 query outcomes were first collapsed across the 48-query mission sequence, Exp.5 was analysed at the mission--closure-event level, and Exp.8 used matched Friedman tests across the five levels of each factor with Kendall's \(W\) and Holm adjustment. Because Exp.1--8 use development maps rather than a separately sampled map population, these inferential quantities characterize within-campaign repeatability, not facility-level generalization.

For the frozen holdout, a run was validated only if it completed before 60 s, respected the resident-node ceiling, returned the required endpoints in order, and passed independent full-segment collision validation. Planner seeds were aggregated within map--mission unit. End-to-end wall time, path length, and clearance used exactly seed-paired common-success outcomes before mission-level aggregation, whereas PAR-2 retained every seed-level failure. Topology-stratified map-cluster bootstrap intervals resampled complete maps within each topology family while carrying all missions belonging to the selected map. Map-level paired effects were tested with two-sided sign-flip permutation tests \cite{ref18}, and Holm correction was applied within endpoint families \cite{ref20}. The archived protocol specified five planner seeds, but the frozen manifest contains three. Consequently, the prespecified 5/5 robust-success endpoint is not testable; seed-averaged mission success and observed 3/3 robust success are reported as secondary endpoints.

The software and analysis pipeline was implemented in Python using NumPy, SciPy, and Matplotlib \cite{ref15,ref16,ref17}.

\section{Results}

\subsection{Development evidence from Exp.1--8}

Across successful development records, every explicitly logged collision-cell and invalid-segment counter was zero. The results below characterize the evaluated development maps and configurations and are kept distinct from the frozen holdout.

\subsubsection{Mission-scale behaviour}

Across the 150 development mission units available at each target count, MIRA-PRM achieved 100\% seed-level success and 100\% all-seed robust success for 2, 4, 6, and 8 ordered targets. Mean total planning time increased from 0.05699 s at two targets to 0.06047, 0.06490, and 0.06938 s at four, six, and eight targets, respectively, corresponding to a 21.7\% increase from two to eight targets. Its mean A*-reference path ratio remained between 1.01334 and 1.01612.

PRM-Rebuild, which reconstructs the roadmap for each segment, increased from 0.07429 to 0.14729, 0.22124, and 0.29625 s, corresponding to a 298.8\% increase from two to eight targets; its robust-success rate decreased from 96.67\% to 84.00\%. Standard reusable PRM remained faster than MIRA-PRM, with mean times of 0.03753--0.04040 s, but produced larger A*-reference ratios of 1.04933--1.05639. MIRA-PRM was faster than PRM* and LE-HG-PRM at every tested target count. Within these three development maps, the principal scaling benefit is therefore relative to repeated construction and denser/guided comparators, not relative to standard PRM.

\subsubsection{Repeated-query adaptation}

Across 12 repeated-query missions, each containing 48 queries and 12 planner seeds, MIRA-PRM achieved 100\% seed-level and robust mission success. Feedback-only, mission-only, and ordinary PRM achieved seed-level success rates of 99.31\%, 97.22\%, and 78.47\%, respectively. Mean total online time was 0.08207 s for MIRA-PRM and 0.08088 s for Feedback-only; their difference of 0.00118 s was not significant after Holm correction (adjusted \(p=0.06397\)). MIRA-PRM nevertheless produced a lower mean reference ratio than Feedback-only (1.09265 versus 1.13134; mean paired difference -0.03963; adjusted \(p=0.00977\)).

For MIRA-PRM, the mean numbers of nodes added at query indices 1--6 were 19.27, 10.78, 9.17, 3.56, 1.74, and 0.48, respectively; no additional nodes were inserted from query 7 to query 48. Thus, refinement became inactive after the early traversals in this campaign. This is empirical stabilization of the configured feedback process rather than a formal convergence result.

\subsubsection{Resident-node budget}

Under the lowest 40-node resident ceiling, MIRA-PRM achieved 97.67\% seed-level success and 76.00\% robust mission success. The corresponding seed-level success rates were 67.17\% for PRM, 86.50\% for PRM*, 83.17\% for Mission-static, and 82.67\% for Structure-PRM. The MIRA-PRM success-rate advantages at this budget were 30.50 percentage points versus PRM and 11.17 points versus PRM*, with Holm-adjusted paired \(p\)-values of $3.49\times 10^{-8}$ and $8.54\times 10^{-6}$, respectively.

MIRA-PRM seed-level success increased to 99.33\% at 70 nodes and reached 100\% at 100, 150, 250, and 400 nodes. Its complete-seed reference ratio decreased from 1.32522 at 40 nodes to 1.03463 at 400 nodes, while mean planning time increased from 0.04553 to 0.06807 s. Ordinary PRM remained faster at every budget, whereas its success increased from 67.17\% to 99.33\% and its reference ratio decreased from 1.58888 to 1.08928. The low-budget experiment therefore shows a reliability--quality gain accompanied by higher computation. HG-PRM and LE-HG-PRM remain contextual per-query comparators because their resident-resource semantics are not matched to the persistent-roadmap methods.

\subsubsection{Target-distribution dependence}

MIRA-PRM, PRM, PRM*, LE-HG-PRM, and A* completed every run under the clustered, cross-zone, alternating, and distributed target patterns; HG-PRM exhibited a small failure rate only in the distributed condition. MIRA-PRM mean planning times were 0.05970, 0.05900, 0.06416, and 0.06559 s for the clustered, cross-zone, alternating, and distributed patterns, respectively. The corresponding reference ratios were 1.01634, 1.01896, 1.02005, and 1.02087.

Across the same patterns, PRM was faster at 0.03354--0.03524 s but had reference ratios of 1.04656--1.06137. MIRA-PRM was faster than PRM*, HG-PRM, and LE-HG-PRM and produced lower reference ratios than all four roadmap comparators in every target-distribution condition; the Holm-adjusted paired path-ratio \(p\)-values were no greater than $3.56\times 10^{-14}$. Within the development cross-zone map, performance was stable across the four tested target arrangements; this should not be generalized to distribution invariance outside those conditions.

\subsubsection{Dynamic local repair}

Both local repair and full rebuilding achieved 100\% success under the gate-lower, gate-upper, and PCS-aisle closure scenarios. Mean event time was 0.01311, 0.01259, and 0.01029 s for local repair, compared with 0.06529, 0.06254, and 0.05693 s for full rebuilding. These differences correspond to event-time reductions of 79.9\%, 79.9\%, and 81.9\%, respectively. Local repair added means of 24.96, 23.96, and 38.69 nodes, compared with 420 nodes for full rebuilding, corresponding to reductions of 94.1\%, 94.3\%, and 90.8\%. The time and node-count contrasts remained significant after Holm correction.

The efficiency gain was not free of geometric cost. The event-level reference ratio of local repair was higher than that of full rebuilding by 0.01523, 0.01264, and 0.01031 in the three closure scenarios. Post-change reference-ratio penalties ranged from 0.00510 to 0.00940, and the PCS-aisle post-change difference was not significant after Holm correction. Local repair therefore provides a clear event-time and graph-growth advantage in these closures, with a small but measurable geometric penalty.

\subsubsection{Mechanism ablation}

In the cumulative ablation, Uniform-PRM achieved 97.50\% seed-level success, 72.00\% robust mission success, a mean planning time of 0.03104 s, and a mean reference ratio of 1.14679. Adding structural sampling increased robust success to 100\% and reduced the reference ratio to 1.09218. Adding inspection support next produced a reference ratio of 1.11428, which was worse than the preceding structure-only stage. Adding mission-flow support reduced the ratio to 1.05370, and the complete MIRA-PRM configuration reduced it further to 1.04190. The complete method required 0.05813 s, 15.44 added nodes, and 1,933 mean collision checks, compared with 0.03104 s, no added nodes, and 1,167 collision checks for Uniform-PRM.

The leave-one-out analysis was also non-monotonic. Removing structure, mission flow, or feedback increased the mean reference ratio from 1.04190 to 1.04563, 1.04859, and 1.05370, respectively. In contrast, removing inspection support reduced the ratio to 1.03808 in this development condition. These paired contrasts were significant after Holm correction. Accordingly, the bottleneck ablation supports the roles of structure, mission flow, and feedback for the evaluated endpoints, while also showing that the inspection-support term is not monotonically beneficial for path ratio. The method should therefore be interpreted as a coupled design rather than as a sum of uniformly positive components.

\subsubsection{Cross-family comparison under configured settings}

Across 90 development map--mission units, MIRA-PRM achieved 100\% seed-level and robust success, a mean planning time of 0.06533 s, and a mean reference ratio of 1.01527. PRM and RRT were faster, with mean times of 0.03826 s and 0.03776 s, but had larger reference ratios of 1.04989 and 1.11373. MIRA-PRM was faster than PRM*, HG-PRM, LE-HG-PRM, MMD-RRT, PRM-Rebuild, and RRT*, while producing lower reference ratios than each of these methods except RRT*. RRT* produced a reference ratio of 1.00147 but required a mean of 12.21445 s.

IAPF achieved only 35.56\% seed-level and robust mission success, and its path-quality summary was available for only 32 of 90 mission units. Its wall-time result should therefore not be interpreted as an efficiency comparison among equally successful planners. Because the planner families use different graph, iteration, collision-check, and termination semantics, Exp.7 characterizes the configured operating settings and cannot be interpreted as a matched-budget ranking.

\subsubsection{One-factor-at-a-time parameter sensitivity}

All 3,200 Exp.8 runs succeeded across the five tested levels of each of the four factors. Across all 20 factor--level conditions, mean planning time ranged from 0.05922 to 0.06202 s, the mean reference ratio ranged from 1.02663 to 1.03594, and the mean number of added nodes ranged from 52.53 to 79.63. The default configuration produced a reference ratio of 1.02848 and approximately 76.88 added nodes.

Friedman tests detected factor-dependent timing differences for all four factors, although the absolute timing range was less than 0.003 s. Path ratio did not differ significantly across the tested uniform-exploration levels after Holm correction (adjusted \(p=0.205\)), but differed across feedback-threshold, mission-mass, and refinement-budget levels (adjusted \(p=0.0126\), \(1.68\times10^{-8}\), and \(0.00288\), respectively). Added-node counts were sensitive to all four factors. Mission completion was stable over the tested one-factor ranges. The sensitivity study does not, however, identify a globally optimal parameter combination or exclude interactions among parameters.

\subsection{Frozen-holdout route reliability}

All 864 expected frozen records were present, with no duplicate keys, invalid timing rows, budget violations, or invalid successful paths. In the 576-run core campaign, validated seed-level success was 96/144 (66.67\%) for MIRA-PRM, 82/144 (56.94\%) for PRM, 89/144 (61.81\%) for PRM*, and 70/144 (48.61\%) for LE-HG-PRM. In the secondary observed-seed analysis, all three seeds succeeded for 30/48 MIRA-PRM missions, compared with 23/48, 27/48, and 19/48 missions for PRM, PRM*, and LE-HG-PRM, respectively.

Observed reliability differed substantially across topology families. Random-obstacle maps reached 100\% success for every method, whereas the 120-node ceiling was insufficient for the maze family. The largest descriptive differences occurred in room-like layouts containing door-like passages and repeated zones, where MIRA-PRM reached 95.83\% seed-level success, compared with 70.83\% for PRM, 85.42\% for PRM*, and 45.83\% for LE-HG-PRM. The concentration of the difference in room-like layouts is consistent with the intended role of geometry and ordered mission-flow guidance, although four maps per family are insufficient to establish a general topology effect.

\begin{table}[t]
\caption{Validated seed-level success by topology; each denominator is 48.}
\label{tab:topology-success}
\centering
\small
\begingroup
\renewcommand{\arraystretch}{1.10}
\begin{tabular}{@{}lcccc@{}}
\hline
\textbf{Map family} & \textbf{MIRA-PRM} & \textbf{PRM} & \textbf{PRM*} & \textbf{LE-HG-PRM} \\
\hline
Maze & 2/48 (4.17\%) & 0/48 & 0/48 & 0/48 \\
Rooms & 46/48 (95.83\%) & 34/48 (70.83\%) & 41/48 (85.42\%) & 22/48 (45.83\%) \\
Random obstacles & 48/48 (100\%) & 48/48 (100\%) & 48/48 (100\%) & 48/48 (100\%) \\
\hline
\end{tabular}
\endgroup
\end{table}

\subsection{Frozen-holdout pairwise effects}

In the secondary map-by-mission analysis of planner-seed success proportions, the estimated success-rate differences were +9.72 percentage points versus PRM, +4.86 points versus PRM*, and +18.06 points versus LE-HG-PRM. The corresponding topology-stratified map-cluster 95\% confidence intervals were [+3.47, +15.28], [+1.39, +9.03], and [+12.50, +26.39] percentage points. However, the Holm-adjusted map-level sign-flip \(p\)-values were 0.250, 0.250, and 0.1875, respectively. These are favourable secondary estimates, but they do not substitute for the unavailable prespecified 5/5 robust-success endpoint.

On common-success map-by-mission medians, MIRA-PRM was 52.23\% slower than ordinary PRM (improvement estimate -52.23\%, 95\% CI [-54.47\%, -45.66\%]). Point estimates indicated 16.89\% shorter end-to-end wall time than PRM* [6.87\%, 22.05\%] and 81.01\% shorter time than the contextual per-query LE-HG-PRM implementation [79.86\%, 82.04\%]. The Holm-adjusted sign-flip \(p\)-values for the PRM* and LE-HG-PRM contrasts were both 0.0625. The corresponding timing effects are therefore reported as supportive point estimates rather than as multiplicity-adjusted evidence of superiority.

Mean PAR-2 time was 40.02 s for MIRA-PRM, 51.68 s for PRM, 45.86 s for PRM*, and 61.75 s for LE-HG-PRM. Relative PAR-2 improvements were 22.56\% [9.07\%, 32.28\%] versus PRM, 12.73\% [3.85\%, 22.21\%] versus PRM*, and 35.19\% [27.35\%, 43.21\%] versus LE-HG-PRM. The PRM* and LE-HG-PRM PAR-2 contrasts remained significant after Holm correction, whereas the PRM comparison remained a favourable but unconfirmed estimate.

MIRA-PRM did not shorten geometric paths. Relative path-length effects were -0.54\% versus PRM, -1.77\% versus PRM*, and -1.86\% versus LE-HG-PRM, where negative values indicate longer MIRA-PRM paths. Their 95\% confidence intervals were [-1.25\%, -0.13\%], [-2.84\%, -1.07\%], and [-2.70\%, -1.02\%], respectively. Every validated route nevertheless had zero occupied-cell crossings, out-of-bounds cells, invalid segments, or goal mismatches. Median minimum-clearance differences were 0.00 cells for all three comparisons, and lower-tail clearance satisfied the prespecified non-inferiority margin.

\begin{table}[t]
\caption{Pairwise effects from the frozen holdout. Positive percentages favour MIRA-PRM.}
\label{tab:pairwise-effects}
\centering
\scriptsize
\begingroup
\renewcommand{\arraystretch}{1.08}
\setlength{\tabcolsep}{3pt}
\begin{tabularx}{\linewidth}{@{}L{0.18\linewidth} C{0.16\linewidth} L{0.23\linewidth} C{0.10\linewidth} X@{}}
\hline
\textbf{Endpoint} &
\textbf{Baseline} &
\textbf{MIRA effect [95\% CI]} &
\textbf{\shortstack{Holm\\adjusted \(p\)}} &
\textbf{Interpretation} \\
\hline
\shortstack[l]{Seed-averaged\\success} & PRM & +9.72 pp [3.47, 15.28] & 0.2500 & Secondary positive estimate \\
\shortstack[l]{Seed-averaged\\success} & PRM* & +4.86 pp [1.39, 9.03] & 0.2500 & Secondary positive estimate \\
\shortstack[l]{Seed-averaged\\success} & LE-HG-PRM & +18.06 pp [12.50, 26.39] & 0.1875 & Secondary positive estimate \\
\shortstack[l]{Observed robust\\3/3} & PRM & +14.58 pp [4.17, 22.92] & 0.5000 & Secondary; not 5/5 \\
\shortstack[l]{Observed robust\\3/3} & PRM* & +6.25 pp [0.00, 12.50] & 0.5000 & Secondary; not 5/5 \\
\shortstack[l]{Observed robust\\3/3} & LE-HG-PRM & +22.92 pp [16.67, 29.17] & 0.3750 & Secondary; not 5/5 \\
\shortstack[l]{Common\\success time} & PRM & -52.23\% [-54.47, -45.66] & 0.0234 & MIRA slower \\
\shortstack[l]{Common\\success time} & PRM* & +16.89\% [6.87, 22.05] & 0.0625 & Supportive estimate \\
\shortstack[l]{Common\\success time} & LE-HG-PRM & +81.01\% [79.86, 82.04] & 0.0625 & Contextual supportive estimate \\
PAR-2 & PRM & +22.56\% [9.07, 32.28] & 0.1250 & Favourable estimate \\
PAR-2 & PRM* & +12.73\% [3.85, 22.21] & 0.0117 & Supported \\
PAR-2 & LE-HG-PRM & +35.19\% [27.35, 43.21] & 0.0117 & Supported \\
Path length & PRM & -0.54\% [-1.25, -0.13] & 0.1250 & Modest length penalty \\
Path length & PRM* & -1.77\% [-2.84, -1.07] & 0.0938 & Modest length penalty \\
Path length & LE-HG-PRM & -1.86\% [-2.70, -1.02] & 0.1250 & Modest length penalty \\
\hline
\end{tabularx}
\endgroup
\end{table}

\begin{figure}[H]
\centering
\includegraphics[width=0.96\linewidth]{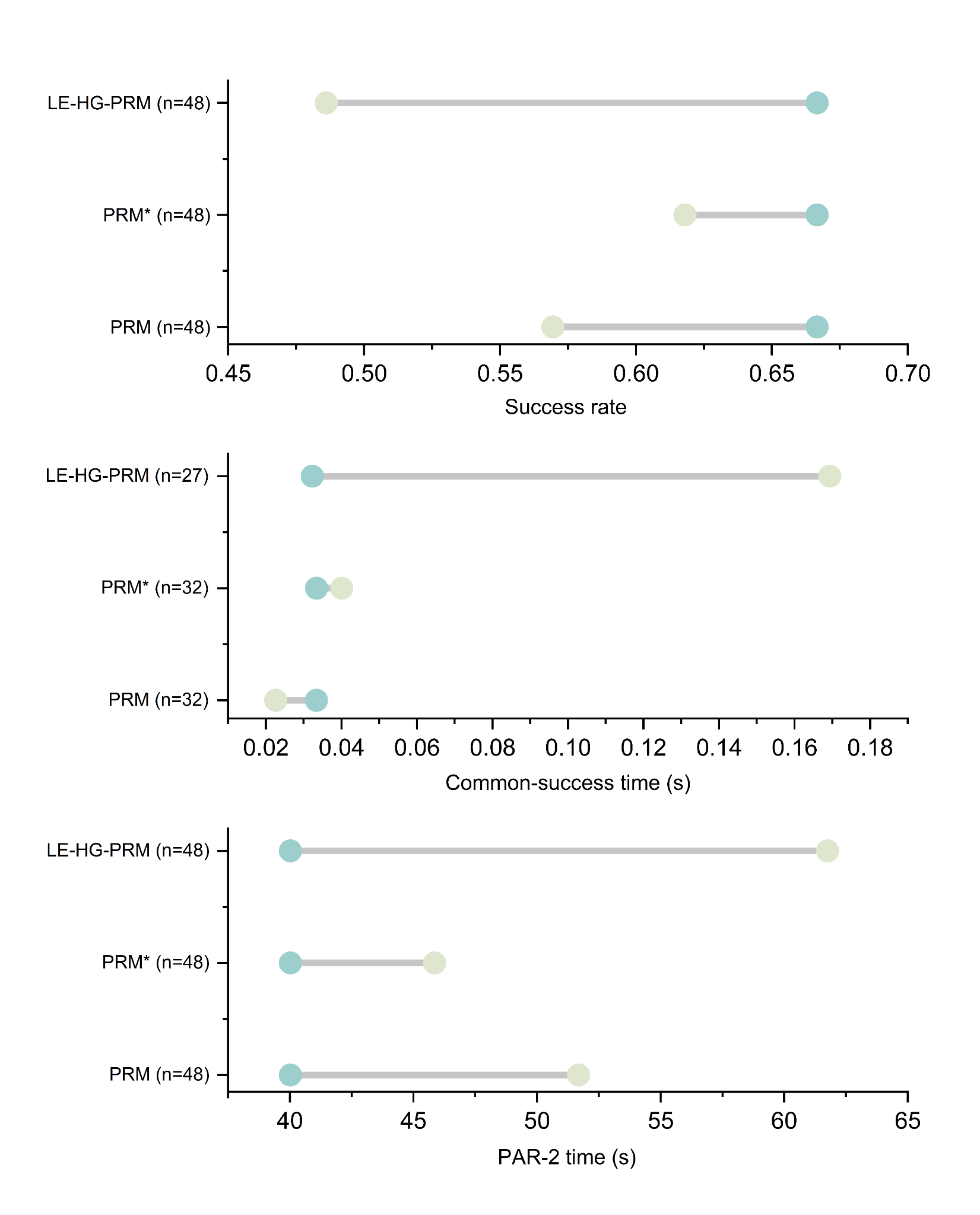}
\caption{Paired outcomes in the frozen holdout (Exp.9). Dots show comparator and MIRA-PRM means over the same map--mission units. Success rate and PAR-2 retain all 48 units per comparison; wall time first uses exactly seed-paired common-success outcomes, takes the within-mission median, and then averages the available missions. Thus, the MIRA-PRM wall-time mean can differ by comparator. \(n\) denotes paired missions. These are descriptive means, not the relative-effect estimates in table~\ref{tab:pairwise-effects}. Cyan dots denote MIRA-PRM and pale-green dots denote the comparator.}
\label{fig:dumbbell}
\end{figure}

\begin{figure}[H]
\centering
\includegraphics[width=0.96\linewidth]{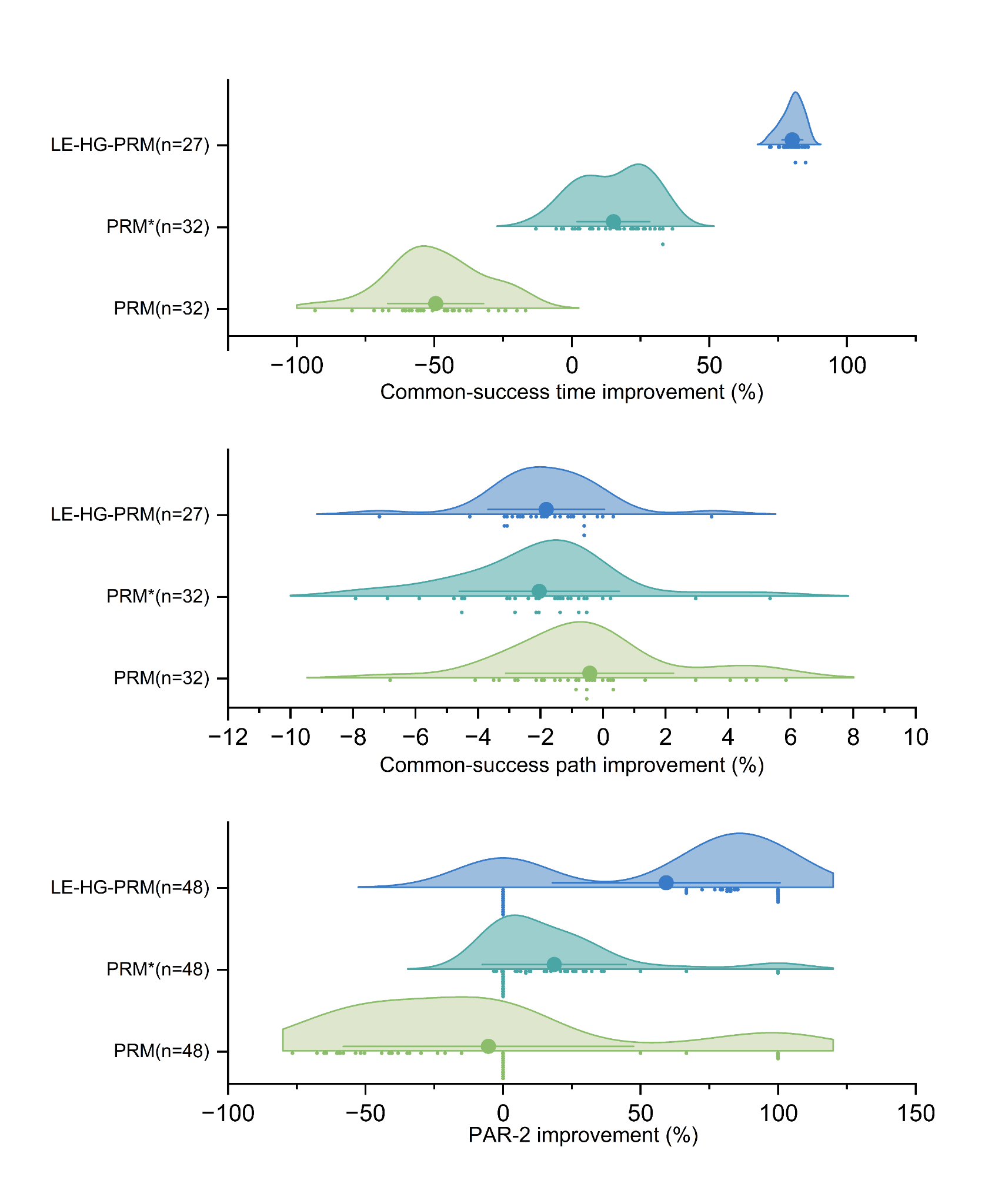}
\caption{Distributions of paired mission-level effects in Exp.9. Each point is one map--mission comparison; half-violins show density and boxes show the median and interquartile range. Positive values indicate lower MIRA-PRM time or path length. Time and path effects use exactly seed-paired common-success outcomes aggregated within mission; PAR-2 includes failures. \(n\) denotes finite paired missions. The observations are descriptive; inferential intervals and tests account for map clustering as reported in table~\ref{tab:pairwise-effects}.}
\label{fig:raincloud}
\end{figure}

Figure~\ref{fig:dumbbell} summarizes paired mission-level outcomes in the frozen holdout. Figure~\ref{fig:raincloud} shows the distributions of paired relative effects. These descriptive displays complement the map-cluster inference in table~\ref{tab:pairwise-effects}; the mission-level observations within a map are not independent map replicates.

\subsection{Resident-node control}

At resident ceilings of 72, 120, and 168 nodes, MIRA-PRM, PRM, and PRM* respected the specified maximum on the frozen resource subset, and every successful path passed the same independent geometric validator. This resource subset verifies resident-node compliance for the tested regime. It does not establish full computational equivalence because collision-check-matched and end-to-end-time-matched regimes were not executed. Peak-memory effects are not reported because that field was not populated in the frozen records.

\section{Discussion}

\subsection{Mechanistic interpretation of the development campaigns}

The development campaigns show where MIRA-PRM gains and costs arise. As the ordered mission length increased, persistent construction avoided the near-linear rebuild cost seen in PRM-Rebuild. Repeated-query refinement became inactive after the early traversals, indicating that the graph retained support useful for later requests. Under low resident-node budgets, the largest reliability difference appeared in the bottleneck layout, where uniform finite sampling was more likely to omit critical passages.

The same campaigns also reveal the costs of the mechanism. Standard PRM and RRT can be faster under successful conditions. Local repair markedly reduces event time and node insertion relative to rebuilding but incurs a small path-quality penalty. The ablation is non-monotonic: removing structure, mission flow, or feedback worsens the mean reference ratio, whereas removing inspection support improves it in the tested bottleneck condition. One-factor sensitivity shows stable mission completion over the evaluated ranges but does not define a unique optimum. These findings favour a coupled reliability--efficiency interpretation rather than a claim that every component improves every metric independently.

\subsection{Topology dependence in the frozen holdout}

The strongest frozen-holdout observation is topology dependent. Room-like layouts contain repeated zones connected through a limited number of door-like passages. With a finite node budget, uniform PRM can fail to represent these gates, whereas PRM* primarily increases graph connectivity. MIRA-PRM instead concentrates sampling probability on constrained geometry and ordered mission flow, while reserving some capacity for route segments that later exhibit weak support. The high success observed in room-like maps is consistent with this mechanism.

The result also has clear bounds. Random-obstacle maps were too easy to separate the methods, while the maze family was too difficult at the 120-node ceiling. Only four maps were available in each topology family. The topology-stratified results should therefore be read as internal evidence about where the proposed mechanism is useful, not as proof of a universal environment-class effect.

\subsection{Implications for mobile gas-sensing measurement}

In mobile gas sensing, an unreachable sampling pose produces missing data regardless of detector sensitivity, selectivity, or calibration. MIRA-PRM targets this route-availability component of the measurement chain. It does not improve the gas detector itself, and the present experiments do not quantify concentration uncertainty. This distinction is important for a measurement-science interpretation: the planner affects whether and when the sensing platform can acquire a scheduled observation, whereas the metrological quality of that observation remains governed by the sensor, sampling protocol, localization, and environmental conditions.

The frozen holdout places MIRA-PRM at an intermediate operating point. It shows higher observed route completion than the three frozen comparators, but a 52.23\% common-success time penalty relative to PRM and an approximately 0.5--1.9\% path-length penalty relative to all three comparators. Conversely, the point estimates favour MIRA-PRM in common-success time relative to PRM* and the contextual per-query LE-HG-PRM implementation, and PAR-2 improves when failures are retained. For a repeated inspection task, the practical value therefore depends on how missing sampling opportunities are weighted relative to additional planning time and modest geometric detour.

\subsection{Limitations}

The present evidence remains simulation based. Exp.1--8 use constructed development maps, so their paired tests quantify repeatability within those maps rather than facility-level generalization. Exp.7 compares configured profiles with different node, collision-check, iteration, and termination semantics and is not a matched-computation ranking. Exp.8 varies one factor at a time and therefore does not resolve parameter interactions. The ablation is tied to a development bottleneck setting and cannot establish independent component effects across arbitrary environments.

The frozen holdout contains 12 procedurally generated maps and no external public facility map set. In addition, the archived protocol specified five planner seeds but the frozen manifest contains three, so the prespecified 5/5 robust-success endpoint is unavailable and the observed 3/3 endpoint remains secondary. Resident-node compliance was verified, but collision-check-matched and total-time-matched resource regimes were not executed, and peak-memory measurements are unavailable. Corridor-width and map-scale stress tests were also not completed.

Finally, MIRA-PRM has not yet been validated on a physical sensing robot. A physical study would need to account for footprint inflation, localization and tracking error, finite pose tolerance, controller dynamics, sensor response time, sampling duration, and gas-measurement uncertainty. Those effects can change both route feasibility and measurement availability and are required before the method can be claimed as a validated mobile measurement solution.

\section{Conclusion}

MIRA-PRM introduces ordered mission information into a persistent probabilistic roadmap through independently normalized geometry, mission-flow, and inspection priors, locally adaptive connectivity, and bounded evidence-gated refinement. Across eight development campaigns, the method showed favourable scaling relative to repeated rebuilding, improved route availability at low resident-node budgets, early stabilization of repeated-query refinement, and efficient local repair. The same experiments also exposed trade-offs: standard PRM and RRT can be faster, local repair carries a small path-quality penalty, and the inspection-support ablation is non-monotonic.

In the separate 12-map frozen holdout, MIRA-PRM produced 96/144 validated successes, compared with 82/144 for PRM, 89/144 for PRM*, and 70/144 for LE-HG-PRM. The largest descriptive reliability difference occurred in room-like layouts. Common-success timing favoured PRM over MIRA-PRM, whereas point estimates favoured MIRA-PRM over PRM* and the contextual per-query LE-HG-PRM implementation; all three path-length comparisons modestly favoured the baselines. Every validated route nevertheless passed independent full-segment checks and the clearance non-inferiority criterion.

Taken together, the results support MIRA-PRM as a reliability-oriented routing strategy for structured repeated-inspection simulations rather than as a universally superior planner. The unavailable prespecified 5/5 endpoint, incomplete resource matching, limited map diversity, and absence of physical gas-sensing validation constrain the claim. External-map replication, matched-computation evaluation, structural-scale testing, and physical mobile-sensing experiments are the next steps required to establish broader measurement-system relevance.

\data{The public implementations of HG-PRM, LE-HG-PRM, and MIRA-PRM are openly available at \url{https://github.com/KEVIN-SyWang/HG-PRM_LE-HG-PRM_MIRA-PRM}. The current public repository is an algorithm-only release and does not contain the internal development and frozen-holdout benchmark/provenance archives used for the numerical analyses in this article. Those supporting data are available from the corresponding authors upon reasonable request because the present public release is limited to algorithm source code. No physical gas-sensing measurements or human-participant data were collected in this study.}

\section*{Conflict of interest}
The authors have no competing interests to declare that are relevant to the content of this article.

\roles{%
Gongsen Wang: Software, Investigation, Writing -- Original Draft, Visualization.\par
Siyuan Wang: Conceptualization, Supervision, Methodology, Investigation, Writing -- Review \& Editing, Visualization.\par
Xinyuan Wang: Methodology, Software, Writing -- Original Draft.\par
Yuhan Wen: Software, Visualization, Writing -- Original Draft.\par
Feng Yang: Software, Writing -- Review \& Editing.\par
Zhen Tian: Data Curation, Investigation, Writing -- Review \& Editing.\par
Hoi Leong Lee: Supervision, Methodology.%
}

\clearpage
\section*{Supplementary material}
\renewcommand{\thetable}{S\arabic{table}}
\renewcommand{\theHtable}{S\arabic{table}}
\setcounter{table}{0}

\begin{table}[H]
\caption{Cross-family comparison under the configured settings in Exp.7.}
\label{tab:configured-comparison}
\centering
\small
\begingroup
\renewcommand{\arraystretch}{1.10}
\setlength{\tabcolsep}{4pt}
\begin{tabularx}{\linewidth}{@{}L{0.19\linewidth} C{0.14\linewidth} C{0.17\linewidth} C{0.14\linewidth} X@{}}
\hline
\textbf{Algorithm} &
\textbf{\shortstack{Success\\(\%)}} &
\textbf{\shortstack{Robust\\success (\%)}} &
\textbf{\mbox{Time (s)}} &
\textbf{\shortstack{A*-reference\\ratio}} \\
\hline
MIRA-PRM & 100.00 & 100.00 & 0.06533 & 1.01527 \\
PRM & 99.86 & 98.89 & 0.03826 & 1.04989 \\
PRM* & 100.00 & 100.00 & 0.08999 & 1.02591 \\
LE-HG-PRM & 100.00 & 100.00 & 0.45061 & 1.02722 \\
HG-PRM & 99.72 & 97.78 & 0.29175 & 1.03670 \\
RRT & 100.00 & 100.00 & 0.03776 & 1.11373 \\
RRT* & 100.00 & 100.00 & 12.21445 & 1.00147 \\
MMD-RRT & 100.00 & 100.00 & 0.08849 & 1.15354 \\
PRM-Rebuild & 99.17 & 93.33 & 0.27860 & 1.04768 \\
A* & 100.00 & 100.00 & 0.05729 & 1.00000 \\
IAPF & 35.56 & 35.56 & 8.28443 & 1.02412 \\
\hline
\end{tabularx}
\endgroup
\end{table}

\noindent\textit{Note.} These comparisons are not resource matched. Path-quality summaries require finite measurements under every configured seed. The IAPF reference ratio is therefore based on 32/90 mission units, whereas the other displayed path-ratio summaries use complete eligible mission sets.

\end{document}